\documentclass[smallextended]{svjour3}
\smartqed

\usepackage{amssymb}
\usepackage{booktabs}
\usepackage{enumitem}
\usepackage{longtable}
\usepackage{graphicx}
\usepackage{array}
\usepackage{dcolumn}
\usepackage{amsfonts}
\usepackage{amssymb}
\usepackage{xspace}
\usepackage{xfrac}
\usepackage{array}
\usepackage{multirow}
\usepackage{xcolor}
\usepackage{footnote}
\usepackage{url}
\usepackage{makeidx}
\usepackage{epsfig}
\usepackage{amsmath}
\usepackage{multicol}
\usepackage{multirow}
\usepackage{mathtools}
\usepackage[misc]{ifsym}
\usepackage[ruled,linesnumbered]{algorithm2e}
\usepackage{tabularx}
\usepackage{array}
\usepackage{footnote}
\makesavenoteenv{tabular}
\makesavenoteenv{table}

\begin{document}

\title{Online Data Augmentation for Forecasting with Deep Learning}

\titlerunning{Online Data Augmentation for Forecasting with Deep Learning}

\author{Vitor~Cerqueira, Moisés~Santos, Luis~Roque, Yassine~Baghoussi, and Carlos~Soares}

\authorrunning{V. Cerqueira et al.}

\institute{Vitor Cerqueira (\Letter) \at
         Faculdade de Engenharia da Universidade do Porto, Porto, Portugal\\
         Laboratory for Artificial Intelligence and Computer Science (LIACC), Portugal\\
         \email{vcerqueira@fe.up.pt}
         \and
         Moisés~Santos \at
         Faculdade de Engenharia da Universidade do Porto, Porto, Portugal\\
         Laboratory for Artificial Intelligence and Computer Science (LIACC), Portugal\\
         \and
         Luis~Roque \at
         Faculdade de Engenharia da Universidade do Porto, Porto, Portugal\\
         Laboratory for Artificial Intelligence and Computer Science (LIACC), Portugal\\
         \and
         Yassine~Baghoussi \at
         Faculdade de Engenharia da Universidade do Porto, Porto, Portugal\\
         INESC TEC, Porto, Portugal\\
         \and
         Carlos~Soares \at
         Faculdade de Engenharia da Universidade do Porto, Porto, Portugal\\
         Laboratory for Artificial Intelligence and Computer Science (LIACC), Portugal\\
         Fraunhofer Portugal AICOS, Portugal\\
         }

\date{Received: date / Accepted: date}

\maketitle


\begin{abstract}

Deep learning approaches are increasingly used to tackle forecasting tasks involving datasets with multiple univariate time series. A key factor in the successful application of these methods is a large enough training sample size, which is not always available.
Synthetic data generation techniques can be applied in these scenarios to augment the dataset.
Data augmentation is typically applied offline before training a model. However, when training with mini-batches, some batches may contain a disproportionate number of synthetic samples that do not align well with the original data characteristics.
This work introduces an online data augmentation framework that generates synthetic samples during the training of neural networks. By creating synthetic samples for each batch alongside their original counterparts, we maintain a balanced representation between real and synthetic data throughout the training process. This approach fits naturally with the iterative nature of neural network training and eliminates the need to store large augmented datasets.
We validated the proposed framework using 3797 time series from 6 benchmark datasets, three neural architectures, and seven synthetic data generation techniques. The experiments suggest that online data augmentation leads to better forecasting performance compared to offline data augmentation or no augmentation approaches. The framework and experiments are publicly available.

\end{abstract}


\section{Introduction}

Time series forecasting is a relevant problem with vast real-world applications. Increasingly, deep neural networks are emerging as effective alternatives to well-established approaches such as ARIMA or exponential smoothing \cite{hyndman2018forecasting}.
Neural architectures such as NHITS \cite{challu2023nhits}, N-BEATS \cite{oreshkin2019n}, or ES-RNN \cite{smyl2020hybrid}, have recently exhibited state-of-the-art forecasting performance in benchmark datasets and competitions.

It is widely accepted that deep neural networks demand substantial data volumes to achieve effective generalization \cite{bandara2021improving}. However, time series datasets often present data scarcity challenges, either having a limited number of time series or containing series with few observations. In effect, a sufficiently large dataset may not be readily available for training deep learning models.

In such data-scarce scenarios, synthetic time series generation techniques can be used to increase the sample size.
Various data generation methods have been developed for time series, ranging from simple approaches such as jittering to more sophisticated techniques involving pattern mixing (averaging multiple time series) \cite{forestier2017generating} or generative models \cite{wen2020time}. 
This work addresses univariate time series forecasting problems with datasets containing multiple time series. In these problems, data augmentation processes have been shown to improve forecasting performance of recurrent-based neural networks \cite{bandara2021improving}. 

In forecasting problems, data augmentation is typically conducted offline before training a model (e.g. \cite{bandara2021improving}). A synthetic dataset is created and combined with the original data, leading to an augmented training set. This augmented dataset is larger and more diverse, thereby improving the performance of models.
However, offline data augmentation has limitations. When training with mini-batches, some batches may contain a disproportionate number of synthetic samples that do not align well with the original data characteristics, potentially misleading the learning process. Additionally, storing and loading a large augmented dataset in memory can be computationally demanding.

To address these limitations, we propose an online data augmentation framework that generates synthetic samples during training. This approach fits naturally with the iterative nature of neural network training, where parameters are updated using mini-batches of data. By creating synthetic samples for each batch alongside their original counterparts, we maintain a balanced representation between real and synthetic data throughout the training process. This balanced representation helps prevent the model from overfitting spurious patterns created during the augmentation process, thereby improving forecasting accuracy. The framework is agnostic to both the augmentation method and neural network architecture, making it widely applicable. Moreover, this approach eliminates the need to store a large augmented dataset by generating synthetic samples online.

While online data augmentation has been explored in speech recognition and computer vision problems \cite{park2019specaugment,nguyen2020improving,hou2023learn}, time series forecasting presents unique challenges due to its temporal nature and the need to preserve complex patterns such as trends and seasonality. Our work is, to our knowledge, the first application of online data augmentation in this context, demonstrating its effectiveness for forecasting tasks.

We validate the proposed method using 3797 time series from 6 benchmark datasets. The results of the experiments suggest that online data augmentation leads to better forecasting performance relative to several offline data augmentation variants or training models without augmentation. 
In summary, the contributions of this work are the following:
\begin{itemize}
    \item A novel framework for online data augmentation for training univariate time series forecasting models using deep learning;

    \item An empirical study comparing the proposed framework across three neural architectures (MLP \cite{hill1996neural}, NHITS \cite{challu2023nhits}, and KAN \cite{han2024kan4tsf}), seven synthetic data generation techniques (from simple transformations to sophisticated pattern mixing approaches), and different training strategies including offline and online data augmentation approaches.

\end{itemize}

All methods are implemented using the neuralforecast Python library, which is based on PyTorch. The experiments are fully reproducible\footnote{\url{https://github.com/vcerqueira/experiments-online_augmentation}} and the framework is available in a Python package\footnote{\url{https://github.com/vcerqueira/metaforecast}}.
The rest of this paper is organized in 5 sections.
Section \ref{sec:2} provides a background to our work, including the definition of the predictive task and an overview of the related work.
The following section (Section \ref{sec:3}) presents the proposed framework for online data augmentation using deep neural networks.
The experiments are presented in Section \ref{sec:4}. We describe the experimental design, including the methods and datasets. We also summarise the results, which are discussed in Section \ref{sec:5}. The conclusions of this work are presented in Section \ref{sec:6}.

\section{Background}\label{sec:2}

This section provides a background to our work.
We start by describing the univariate time series forecasting problem in Section \ref{sec:2.1}. Then, we explore how deep learning is used to tackle this task, focusing on how these methods leverage datasets with multiple time series to train a model (Section \ref{sec:2.2}). Finally, in Section \ref{sec:2.3}, we overview the related work on data augmentation in the context of time series.

\subsection{Univariate Time Series Forecasting}\label{sec:2.1}

A univariate time series is a time-ordered sequence of values $Y = \{y_1, y_2, \dots,$ $y_t \}$, where $y_i \in \mathbb{R}$ is the value of $Y$ observed at time $i$ and $t$ is the length of $Y$.
Forecasting is the process of predicting the value of upcoming observations $y_{t+1}, \ldots, y_{t+h}$, where $h$ denotes the forecasting horizon. 
The importance of forecasting spans several domains, including applications in business areas such as inventory management or operations planning \cite{kahn2003measure}.

Machine learning methods tackle forecasting problems  using an auto-regressive type of modeling. According to this approach, each observation of a time series is modeled as a function of its past $q$ lags based on time delay embedding~\cite{bontempi2013machine}. 
Time delay embedding is the process of transforming a time series from a sequence of values into an Euclidean space. In practice, the idea is to apply sliding windows to build a dataset $\mathcal{D}=\{$X$, $y$\}^t_{q+1}$ where $y_i$ represents the $i$-th observation and $X_i \in \mathbb{R}^q$ is the $i$-th corresponding set of $q$ lags: $X_i = \{y_{i-1}, y_{i-2}, \dots, y_{i-q} \}$. Accordingly, the objective is to train a regression model to learn the dependency $y_i = f(X_i)$. 

Forecasting problems often involve datasets with multiple univariate time series. 
We define a collection of $n$ time series as $\mathcal{Y} = \{Y_1, Y_2, \dots, Y_n\}$. 
Traditional approaches build a forecasting model for each univariate time series within a collection $\mathcal{Y}$. Such models are commonly known as local models \cite{januschowski2020criteria}.
Leveraging the data of all available time series can be valuable to build a forecasting model. The dynamics of the time series within a collection are often related, and a model may learn useful patterns in some time series that are not revealed in others. Models that are trained on collections of time series are referred to as global forecasting models \cite{godahewa2021ensembles}.

For global approaches, the time delay embedding framework described above is applied to each time series in the collection.
Then, an auto-regression model is trained on the joint dataset that contains all transformed time series.
In effect, for a collection of time series the dataset $\mathcal{D}$ is composed of a concatenation of the individual datasets: $\mathcal{D} = \{\mathcal{D}_1, \dots,  \mathcal{D}_n\}$, where $\mathcal{D}_j$ is the dataset corresponding to the time series $Y_j$.

\subsection{Forecasting with Deep Learning}\label{sec:2.2}

In most cases, global forecasting models are trained using deep neural networks. Various types of neural architectures have been recently developed for time series forecasting.
Architectures based on recurrent neural networks, such as the LSTM \cite{siami2018comparison}, are more common due to their intrinsic capabilities for sequence modeling.
However, deep neural networks based on convolutional layers \cite{koprinska2018convolutional}, transformers \cite{zhou2021informer,lim2021temporal}, or multi-layer perceptrons (MLPs) \cite{challu2023nhits} have also shown competitive forecasting accuracy.

The adoption of deep learning to solve forecasting problems surged when Smyl \cite{smyl2020hybrid} presented an LSTM-based model that won the M4 forecasting competition that featured 100.000 time series from various application domains. The method, known as ES-RNN, combines exponential smoothing with an LSTM neural network. Exponential smoothing is used to de-seasonalize and normalized time series, while the neural network is trained on the transformed data in a global fashion. Another popular recurrent-based architecture is DeepAR \cite{salinas2020deepar}. DeepAR leverages a stack of auto-regressive LSTM layers and Markov Chain Monte Carlo sampling to produce probabilistic forecasts.

Several deep learning forecasting methods adopt the transformer architecture, which has been driving important advances in natural language processing tasks. Examples include the Temporal Fusion Transformer \cite{lim2021temporal}, and the Informer \cite{zhou2021informer}. However, recent works have undermined the effectiveness of transformer-based methods for forecasting problems \cite{zeng2023transformers}.

The MLP is one of the simplest neural network architectures, which is composed of stacks of fully connected layers. Neural networks based on MLPs have been used to tackle forecasting problems for several years \cite{hill1996neural}. Recently, two particular architectures have shown promising forecasting accuracy and exceptional computational efficiency: N-BEATS \cite{oreshkin2019n} and NHITS \cite{challu2023nhits}.
Both are based on stacks that contain blocks of MLPs along with residual connections.
NHITS \cite{challu2023nhits}, short for Neural Hierarchical Interpolation for Time Series Forecasting, extends N-BEATS \cite{oreshkin2019n} by including multi-rate input sampling that models data with different scales and hierarchical interpolation for better long-horizon forecasting.
NHITS has shown state-of-the-art forecasting performance relative to other deep learning forecasting methods, including several transformers, and recurrent-based neural networks \cite{challu2023nhits}.
Challu et al. \cite{challu2023nhits} also provide evidence that NHITS is more computationally efficient than transformer-based methods by a factor of 50. 

Kolmogorov-Arnold Network (KAN)~\cite{liu2024kan} have been recently developed as an alternative to MLPs by featuring learnable
activation functions on the weights of neural networks. KANs are inspired by the Kolmogorov-Arnold Representation theorem, providing theoretical guarantees regarding function approximation. Concerning time series forecasting, recent works \cite{han2024kan4tsf,vaca2024kolmogorov} have compared KANs with different neural networks, including transformers, MLPs, or convolutional-based neural networks.

\subsection{Time Series Data Augmentation}\label{sec:2.3}

One key motivation for adopting a global approach is the additional data available for training a model. Machine learning algorithms tend to perform better with larger training sets \cite{cerqueira2022case}. 
However, an adequate sample size may not be readily available. This issue can be solved using data augmentation techniques \cite{bandara2021improving}. 

\subsubsection{Generation methods}

There are several methods for time series data augmentation. Flipping or jittering the value of observations are among the simplest approaches to create a synthetic time series from an original one \cite{wen2020time,roque2024rhiots}. Jittering involves adding random noise to each data point in the original time series with the goal of making forecasting models robust to such noise \cite{um2017data}. The noise can be defined as $\varepsilon \sim \mathcal{N}(0, \sigma'^2)$, where $\sigma'$ is the standard deviation of the input time series scaled by some value $s$ ($\sigma' = \sigma \times s$).

Scaling \cite{wen2020time} is another time series transformation technique that creates synthetic samples by multiplying each observation by a random factor drawn from a normal distribution centered at 1.0, specifically $\mathcal{N}(1, \sigma^2)$. Scaling affects the magnitude of the time series -- a factor below 1 decreases amplitude while a factor above 1 increases it.

Time warping \cite{rashid2019window} is a technique that distorts the time variable of times series to create synthetic samples. The method works by scaling knot points along the time axis using random factors sampled from a $\mathcal{N}(1, \sigma^2)$, thus creating a smooth warping function using cubic splines. Then, the time series is resampled based on these distortions. In an activity recognition task, Rashid and Louis \cite{rashid2019window} show that time warping improves the performance of several machine learning models, including neural networks.

Magnitude warping \cite{iwana2021empirical} is similar to time warping but applied directly to the values of the time series.  
The idea is to apply distortions on a cubic spline to the values of time series. Magnitude warping is also related to scaling. The main difference is that scaling produces local changes that can vary abruptly between consecutive observations, while the changes produced by magnitude warping are smooth due to the spline interpolation.

\begin{figure}[bh]
    \centering
    \includegraphics[width=.9\textwidth, trim=0cm 0cm 0cm 0cm, clip=TRUE]{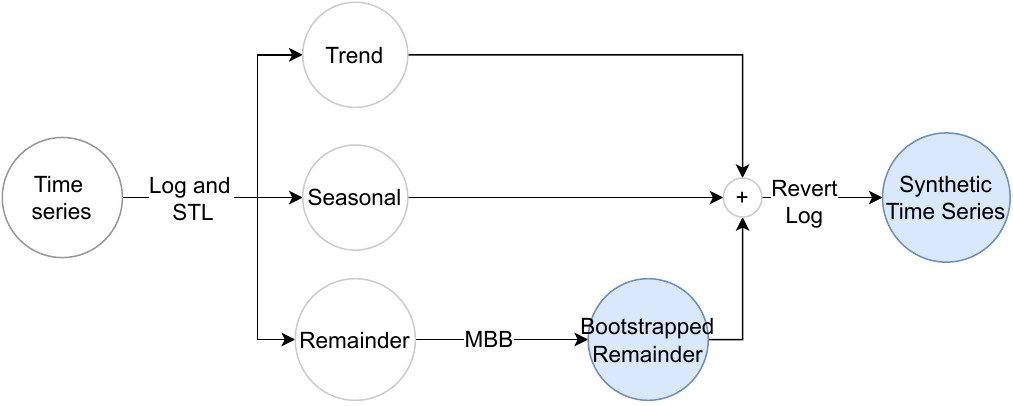}
    \caption{High-level workflow of the synthetic time series generation based on STL and MBB. The blue shaded circle represent synthetic data.}
    \label{fig:mbb_workflow}
\end{figure}

Jittering, scaling, time warping, and magnitude warping are data generation methods based on applying transformation to the original samples. There are several techniques that aims to mix the patterns of one or more time series to generate new ones. One of these techniques is moving blocks bootstrapping (MBB) \cite{lahiri2013resampling} coupled with seasonal decomposition \cite{cleveland1990stl}. Following 
Bergmeir et al. \cite{bergmeir2016bagging}, the idea is to extract the remainder component of a given log-transformed univariate time series, and resample it using MBB. The resampled remainder is then combined back with the other components, leading to a bootstrapped version of the original series (c.f. Figure \ref{fig:mbb_workflow}). This approach has been shown to improve forecasting performance both in terms of offline data augmentation \cite{bandara2021improving} as well as bagging forecasting models \cite{bergmeir2016bagging}. 

DBA \cite{forestier2017generating}, short for DTW (dynamic time warping) Barycentric Averaging, is another technique based on pattern mixing that involves averaging multiple time series to create a synthetic one. DBA works by first selecting a subset of time series from a given dataset. Then, these are combined using a weighted averaged based on DTW where the weights are sampled from a Dirichlet distribution.

TSMixup \cite{ansari2024chronos} is another time series synthetic generation method that combines the patterns of multiple time series. The method randomly samples a subset of time series of a specified length, and then combines them using weights drawn from a Dirichlet distribution. TSMixup resembles time series morphing \cite{santos2023tsmorph}, which consists in gradually transforming a time series into another.

\begin{figure}[ht]
    \centering
    \includegraphics[width=\textwidth, trim=0cm 0cm 0cm 0cm, clip=TRUE]{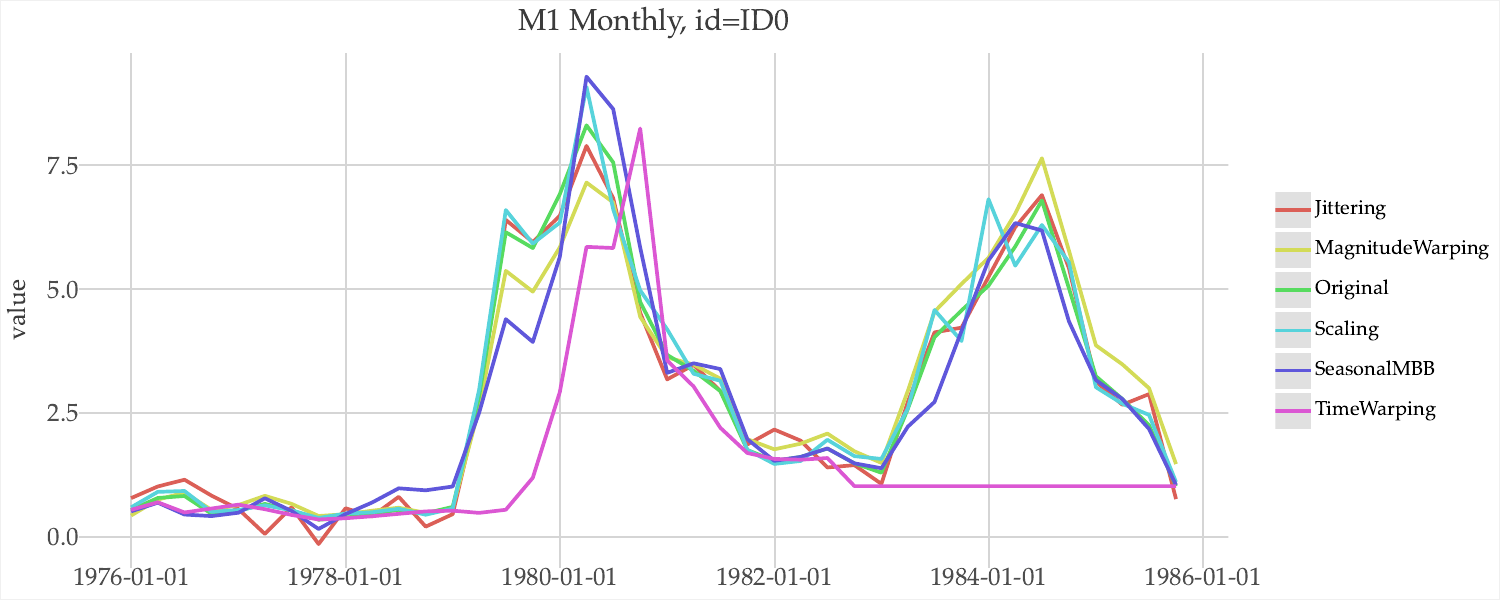}
    \caption{Example of the synthetic time series generated from an original one. This example focus on transformations that create a synthetic time series without reference to other series.}
    \label{fig:example_aug}
\end{figure}

\noindent Figure~\ref{fig:example_aug} shows an example of the application of several transformations to a given a time series (time series  with id \texttt{ID0} from the M1 monthly dataset). This example focus on transformations that create a synthetic time series without reference to other series. This excludes methods such as TSMixup or DBA which combine the patterns of multiple time series to generate synthetic ones.

\subsubsection{Augmentation approach}

\begin{figure}[ht]
    \centering
    \includegraphics[width=.85\textwidth, trim=0cm 0cm 0cm 0cm, clip=TRUE]{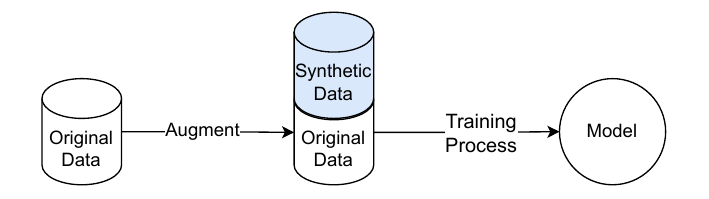}
    \caption{Typical workflow for time series data augmentation in the context of forecasting. A set of synthetic time series is created and concatenated with the original data, and the combined dataset is used to train a forecasting model.}
    \label{fig:apriori}
\end{figure}

The synthetic time series data generation techniques described above are typically used to augment training datasets and to build more accurate models. For example, Bandara et al. \cite{bandara2021improving} tested several of these methods for data augmentation and reported improvements in forecasting accuracy with an LSTM neural network. The augmentation process is conducted before the training process. As illustrated in Figure \ref{fig:apriori}, they first augment the original time series dataset using one of these methods and then train an LSTM on the augmented dataset. Concerning the number of time series generated for a given dataset, Bandara et al. \cite{bandara2021improving} use an heuristic based on the number of time series available. 

\section{Methodology}\label{sec:3}

This section describes the methodology for training a univariate time series forecasting model based on artificial neural networks using online data augmentation. We start by describing the data and main preprocessing steps (Section \ref{sec:3.1}), and then detail the online data augmentation framework (Section \ref{sec:3.2}).

\subsection{Pre-processing Steps}\label{sec:3.1}

We address univariate time series multi-step forecasting problems using artificial neural networks, focusing on datasets that contain multiple time series. This type of dataset may be composed of individual time series with distinct scales or variances. We pre-process each time series using standardization to stabilize the variance of the dataset and bring the data into a common scale. Standardization is also beneficial for improving the efficiency of the training of neural networks by improving gradient descent convergence and preventing neuron saturation \cite{goodfellow2016deep}. To standardize a given time series $Y$, we subtract the mean of the time series and divide by its standard deviation:

\begin{equation}
    y'_i = \frac{y_i - Y_{\mu}}{Y_{\sigma}}
\end{equation}

\noindent where $Y_{\mu}$ is the mean of the time series and $Y_{\sigma}$ the standard deviation. This transforms Y into $Y' = {y'_1, y'_2, \dots, y'_t}$ with zero mean and unit variance. We note that the standardization parameters ($Y_{\mu}$ and $Y_{\sigma}$) are computed using the training data. For forecasting, we first obtain predictions in the standardized scale and then reverse the transformation by multiplying by $Y_{\sigma}$ and adding $Y_{\mu}$.

Having standardized the individual time series, we apply the auto-regressive framework described in Section \ref{sec:2.1} to build a global forecasting models using a neural network. Given the training set that contains a collection of time series $\mathcal{Y}$, we first standardize each time series independently. Then, we apply time delay embedding using a window of size $q$ to transform each standardized series into a dataset ready for supervised learning. At each timestep, the following $h$ (forecasting horizon) observations are modeled based on the recent past $q$ values. The resulting samples are concatenated to form a single training set $\mathcal{D}$, which is used to train a neural network. The training procedure follows a multi-input multi-out strategy for multi-step forecasting \cite{taieb2010multiple}, since neural networks can naturally model multiple output variables.

\subsection{Online Data Augmentation}\label{sec:3.2}

Neural networks are inherently online, with their parameters being updated iteratively based on mini-batches of data. The proposed training framework leverages this property by conducting data augmentation online, i.e., during the training process. 

\begin{figure}[ht]
    \centering
    \includegraphics[width=.95\textwidth, trim=0cm 0cm 0cm 0cm, clip=TRUE]{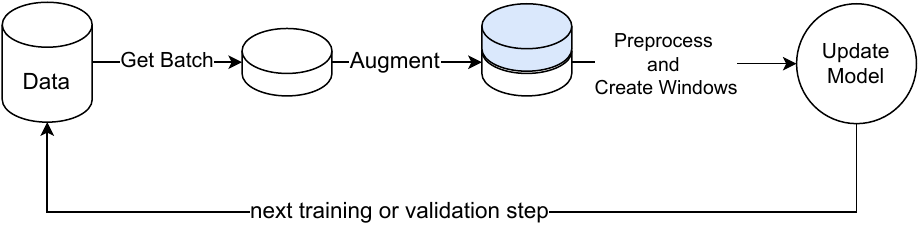}
    \caption{High-level workflow of the training process of a neural network using the proposed framework for online data augmentation.}
    \label{fig:otf_scheme}
\end{figure}

Figure \ref{fig:otf_scheme} shows a high-level diagram that details the workflow in a given training step of a neural network for univariate time series forecasting. 
First, we get a batch of time series, following the standard mini-batch approach for training neural networks. We augment this batch using some time series synthetic data generation technique. The framework is agnostic to the underlying generation technique, and we test several ones in the experiments. In general, the generation method should be both 1) non-deterministic, so different variations are created at each run on the same time series; and 2) computationally efficient, so it is feasible to apply the method at each training or validation step of a neural network. On top of these, our framework also assumes that the underlying temporal patterns of the original time series are preserved. In other words, the synthetic time series should maintain the essential characteristics of the data while providing useful alternative representations for learning.

For each time series in a given batch, we create a single synthetic variation of it, doubling the number of time series in the batch.
We hypothesize that a one-to-one ratio provides a balanced representation between original and synthetic samples. This helps the model learn from real data while benefiting from the additional patterns and variations introduced by data augmentation. Moreover, not creating too many synthetic time series may prevent the model from overfitting to artificial patterns.

After augmentation, data pre-processing is conducted based on the techniques described on the previous section (Section \ref{sec:3.1}). The processed data is then used to update the model parameters. Note that both original and augmented data are used to compute the loss and update neural network parameters. After this, the synthetic data created for this batch is discarded, and the process is repeated until the training process finishes.

\section{Experiments}\label{sec:4}

This section presents the experiments conducted to analyse the performance of online data augmentation. The central research question posed in these experiments is the following: how does online data augmentation perform relative to other training approaches for univariate time series forecasting based on deep learning? This question is addressed based on forecasting accuracy computed from various perspectives.

\subsection{Data}\label{sec:datasets}

We use 6 datasets from four databases and forecasting competitions: M1 \cite{makridakis1982accuracy}, M3 \cite{makridakis2000m3}, and Tourism \cite{athanasopoulos2011tourism}.
M1, and M3 are time series databases that come from the Makridakis competitions. These datasets cover different application domains, including industry, demography, and economics. Tourism is a database focused on the tourism domain.
These datasets, which represent standard benchmarks for univariate time series forecasting, are summarised in Table~\ref{tab:data}. 

\begin{table}
\caption{Summary of the datasets: average value, number of time series, number of observations, seasonal period, and forecasting horizon.}
\label{tab:data}
\resizebox{.95\textwidth}{!}{ %
\begin{tabular}{lrrrrr}
\toprule
Dataset & Average value & \# time series & \# observations & Period & h \\
\midrule
M1 Monthly & 72.7 & 617 & 44892 & 12 & 12 \\
M1 Quarterly & 40.9 & 203 & 8320 & 4 & 8 \\
M3 Monthly & 117.3 & 1428 & 167562 & 12 & 12 \\
M3 Quarterly & 48.9 & 756 & 37004 & 4 & 8 \\
Tourism Monthly & 298.5 & 366 & 109280 & 12 & 12 \\
Tourism Quarterly & 99.6 & 427 & 42544 & 4 & 8 \\
\midrule
Total & - & 3797 & 409602 & - & - \\
\bottomrule
\end{tabular}%
}
\end{table}

We focus on databases with low-frequency time series, namely monthly and quarterly time series. 
These are the ones that tend to comprise a lower number of observations, thus where data augmentation can be more useful. Overall, the combined datasets contain 409602 observations across 3797 univariate time series.
In terms of forecasting horizon, we set this value to 12 and 8, for monthly and quarterly time series, respectively. The input size (number of lags) is set to  24 for monthly time series and 8 for quarterly ones. These values correspond to two seasonal periods, which provide a robust setup according to previous work \cite{leites2024lag}.

\subsection{Evaluation}\label{sec:exp_splits}

\begin{figure}[ht]
    \centering
    \includegraphics[width=.75\textwidth, trim=0cm 0cm 0cm 0cm, clip=TRUE]{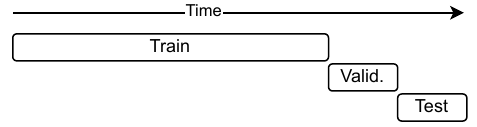}
    \caption{Traditional time series data partitioning process into training, validation, and test sets for developing forecasting models.}
    \label{fig:trad_partitioning}
\end{figure}

For performance estimation, we leave the last $h$ observations of each time series for testing. The remaining available observations are used for training and validating the model. The validation set is composed of the final $h$ observations of each time series, similarly to the test split. The data partitioning process is illustrated in Figure \ref{fig:trad_partitioning}.

We use the Mean Absolute Scaled Error (MASE) as evaluation metric, which is defined as follows:

\begin{equation}
    \text{MASE} = \frac{\frac{1}{n}\sum_{i=1}^{n}|y_i - \hat{y}_i|}{\frac{1}{n-m}\sum_{i=m+1}^{n}|y_i - y_{t-m}|}
\end{equation}

\noindent where $\hat{y}_i$, and $y_i$ are the forecast and actual value for the $i$-th instance, respectively. MASE is a scale-independent measure of forecast accuracy. The mean absolute error of a given model, computed across $n$ data points, is scaled by the mean absolute error of a seasonal naive forecast with period $m$.

\subsection{Methods}\label{sec:exp_methods}

This section describes the methods used in the experiments. First, we present the synthetic time series generation methods used for data augmentation (Section \ref{sec:methods1}). In Section \ref{sec:methods2}, we described the different approaches used to augment time series datasets based on those time series generation methods. Finally, we also detail the neural networks used to train forecasting models (Section \ref{sec:methods3}).

\subsubsection{Synthetic time series generation methods}\label{sec:methods1}

We evaluate the proposed approach using 7 different time series data generation approaches, specifically:

\begin{itemize}
    \item Seasonal MBB (\texttt{MBB}): The seasonal MBB method based on an STL decomposition and bootstrapping of residuals (c.f. Figure~\ref{fig:mbb_workflow}). By default, this technique includes a log transformation for variance stabilization;
    \item \texttt{Jittering}: Adding random Gaussian noise to each observation in the time series. The main parameter of this technique is $s$, the proportion of the time series standard deviation that controls the jittering magnitude (c.f. Section \ref{sec:2.3});
    \item \texttt{Scaling}: Multiplying each observation by random factors drawn from a normal distribution centered at 1.0 and a standard deviation $\sigma$;
    \item Time Warping (\texttt{T-Warp}): Creating synthetic samples by distorting the time axis using cubic splines, including scaling knot points using factors randomly sampled from a $\mathcal{N}(1, \sigma^2)$;
    \item Magnitude Warping (\texttt{M-Warp}): Applying smooth distortions to the values of time series using cubic splines, also including scaling knot points using factors randomly sampled from a $\mathcal{N}(1, \sigma^2)$;
    \item \texttt{TSMixup}: Combining a subset of time series using weights sampled from a Dirichlet distribution. At each run, the number of time series to be combined is randomly sampled from 2 to 10;
    \item \texttt{DBA}: Averaging a subset of time series using DTW and weights sampled from a Dirichlet distribution. Similarly to \texttt{TSMixup}, we sample a number of time series from 2 to 10 to average.
\end{itemize}

\noindent We refer to Section \ref{sec:2.3} for complete description of these methods. Table~\ref{tab:methods_params} shows the parameters of each of these approaches. The parameter values in bold are used by default. In some cases,  which will be described in the next section (Section \ref{sec:methods2}), other configurations are sampled from the list.

\begin{table}[ht]
    \centering
    \caption{Parameters of the time series synthetic generation methods.}
    \label{tab:methods_params}
    \resizebox{.95\textwidth}{!}{ %
    \begin{tabular}{lll}
        \toprule
        Method & Parameter & Values \\
        \midrule
        \texttt{MBB} & log & \{\textbf{True}, False\} \\
        \midrule
        \texttt{Jittering} & $s$ & \{\textbf{0.03}, 0.05, 0.1, 0.15, 0.2, 0.3\} \\
        \midrule
        \texttt{Scaling} & $\sigma$ scaling factor in $\mathcal{N}(1, \sigma^2)$ & \{0.03, 0.05, \textbf{0.1}, 0.15, 0.2, 0.3\} \\
        \midrule
        \multirow{2}{*}{\texttt{M-Warp}} & $\sigma$ scaling factor in $\mathcal{N}(1, \sigma^2)$ & \{0.05, \textbf{0.1}, 0.15\} \\
         & \# knots & \{3, \textbf{4}, 5\} \\
        \midrule
        \multirow{2}{*}{\texttt{T-Warp}} & $\sigma$ scaling factor in $\mathcal{N}(1, \sigma^2)$ & \{0.05, \textbf{0.1}, 0.15\} \\
         & \# knots & \{3, \textbf{4}, 5\} \\
        \midrule
        \multirow{2}{*}{\texttt{DBA}} & Max \# time series & \{5, 7, \textbf{10}, 15\} \\
         & Dirichlet concentration & \{\textbf{1.0}, 1.5, 2.0\} \\
        \midrule
        \multirow{2}{*}{\texttt{TSMixup}} & Max \# time series & \{5, 7, \textbf{10}, 15\} \\
         & Dirichlet concentration & \{\textbf{1.0}, 1.5, 2.0\} \\
        \bottomrule
    \end{tabular}%
    }
\end{table}

\subsubsection{Augmentation approaches}\label{sec:methods2}

Each of the methods described in the previous section are used for augmenting the training dataset that is used to build a forecasting model. We use the following 6 data augmentation strategies:

\begin{itemize}
    \item \texttt{Online}: The proposed online data augmentation scheme (c.f. Figure~\ref{fig:otf_scheme}), in which each batch of time series in a given training step is augmented. We use a batch size (for the original set of time series) of 32 and create a synthetic time series for each one of these. This means that, for training, the batch size is 64 where half of them are synthetic. The parameters of the respective synthetic data generation method are fixed based on Table~\ref{tab:methods_params}.

    \item \texttt{Online(E)}: A variant of \texttt{Online} where the parameters of the synthetic data generation method are randomly sampled in each batch. The possible configurations are described in Table~\ref{tab:methods_params};

    \item \texttt{Offline(1)}: An approach that does data augmentation before the fitting process, following Bandara et al. \cite{bandara2021improving}. First, 1 synthetic time series is created for each one available in the original training dataset. Then, a model is trained using the augmented data. The batch size in this case is set to 64 to match the value used in the \texttt{Online} variants;

     \item \texttt{Offline(10)}: A variant of \texttt{Offline(1)}, but creating 10 synthetic time series for each one available on the training set, following the heuristic described by Bandara et al. \cite{bandara2021improving};

    \item \texttt{Offline(=)}: Another variant of \texttt{Offline(1)}, but creating a number of synthetic time series to match the synthetic sample size created by the \texttt{Online} approach. When applying the \texttt{Online} data augmentation strategy following the parameters described before, we create a total of 32000 time series (32 batch size times 1000 training steps). To compute the number of synthetic time series required to match the sample size of \texttt{Online}, we divide this value (32000) by the number of time series in the dataset.

    \item \texttt{Offline(=, E)}: A variant of \texttt{Offline(=)}, but varying the parameters of the time series synthetic data generation method. Specifically, each time a new time series is created, we randomly sample the generation method parameters based on the configuration pool described in Table~\ref{tab:methods_params}.

\end{itemize}

Besides these, we also include an \texttt{Original} training procedure that does not involve data augmentation.

\subsubsection{Neural network architectures and baseline}\label{sec:methods3}

The time series augmentation approaches described above are tested with three different neural networks: NHITS~\cite{challu2023nhits}, MLP \cite{hill1996neural}, and KAN \cite{han2024kan4tsf}. These are briefly described in Section~\ref{sec:2.2}. These particular neural networks have shown competitive forecasting accuracy relative to other deep learning approaches, including several transformers or recurrent-based neural networks \cite{challu2023nhits,zeng2023transformers}. On top of their competitive forecasting accuracy, these methods are also more computational efficient than other popular architectures \cite{challu2023nhits} (c.f. Section~\ref{sec:2.2}).
For all three neural networks, we use the implementation available on neuralforecast\footnote{\url{https://nixtlaverse.nixtla.io/neuralforecast/models.html}} Python library, which is based on PyTorch. 
In terms of training protocol, we build one global forecasting model for each dataset listed in Table \ref{tab:data}. Besides these neural networks, we also include the seasonal naive method (\texttt{Naive}) in the experiments. The baseline technique that uses the last known observation of the same season as the forecast.

We fix the configuration of the architectures for all training procedures. In all three cases, we adopt the default configuration. These can be described as follows: 
\begin{itemize}
    \item NHITS: Following Challu et al. \cite{challu2023nhits}, each NHITS model is composed of 3 stacks, each of which with one block of MLPs. Each MLP contains 2 hidden layers, each with 512  units. The activation function is set to the rectified linear unit (ReLU). The initial learning rate is 0.001, which is updated 3 times during training. 

    \item MLP: A densely-connected architecture with 2 hidden layers, each with 1024 units. The activation function is ReLU and the learning rate is set to 0.001;

    \item KAN: A KAN neural network with a single hidden layers composed of 512 units. The order of the splines is 3.
\end{itemize}

\noindent In all three methods, the training process run for a maximum of 1000 steps using the ADAM optimizer. For preprocessing, in all cases the time series are standardized.
The three neural networks are trained in an auto-regressive manner with an input size corresponding to two seasonal cycles (c.f. Section \ref{sec:datasets}).

\subsection{Results}

This section presents the results of the experiments. Table~\ref{tab:scores_by_ds} shows the average MASE of each data augmentation approach (columns) when applied with different neural networks and time series generation methods (rows), across the 6 datasets listed in Table \ref{tab:data}. The bold and underlined result represents the best and second-best score in the respective (neural network, synthetic data generation method) pair.

\begin{table}[!h]
\caption{Average forecasting accuracy (MASE) of each data augmentation approach (columns) when applied with different neural networks and time series generation methods (rows), computed across all 6 datasets. The bold and underlined result represents the best and second-best score in the respective (neural network, synthetic data generation method) pair. To facilitate visual comparison, we repeat the results of \texttt{Original} and \texttt{Naive} in each row, though no data augmentation is conducted in those cases.}
\label{tab:scores_by_ds}
\resizebox{.95\textwidth}{!}{ %
\begin{tabular}{llllllll||ll}
\toprule
 &  & \rotatebox{60}{Online} & \rotatebox{60}{Online(E)} & \rotatebox{60}{Offline(1)} & \rotatebox{60}{Offline(10)} & \rotatebox{60}{Offline(=)} & \rotatebox{60}{Offline(=,E)} & \rotatebox{60}{Original} & \rotatebox{60}{Naive} \\
\midrule
\multirow{7}{*}{\hspace{-3pt}\rotatebox{90}{\makebox[0pt]{KAN}}} & DBA & \textbf{1.0886} & \underline{1.0904} & 1.0938 & 1.0932 & 1.0977 & 1.7433 & 1.1312 & 1.3427 \\
 & Jittering & \underline{1.0947} & \textbf{1.0739} & 1.1179 & 1.111 & 1.1065 & 1.674 & 1.1312 & 1.3427 \\
 & M-Warp & \underline{1.0945} & \textbf{1.0922} & 1.108 & 1.1251 & 1.1258 & 1.6714 & 1.1312 & 1.3427 \\
 & Scaling & \underline{1.0804} & \textbf{1.0778} & 1.1 & 1.1039 & 1.1304 & 1.6913 & 1.1312 & 1.3427 \\
 & MBB & \underline{1.0912} & \textbf{1.0837} & 1.1055 & 1.1044 & 1.1097 & 1.6935 & 1.1312 & 1.3427 \\
 & TSMixup & \textbf{1.088} & \underline{1.0926} & 1.1239 & 1.1466 & 1.1213 & 1.6829 & 1.1312 & 1.3427 \\
 & T-Warp & \textbf{1.1105} & \underline{1.111} & 1.1294 & 1.2637 & 1.4995 & 1.736 & 1.1312 & 1.3427 \\
\noalign{\smallskip}
\cline{1-10}
\noalign{\smallskip}
\multirow{7}{*}{\hspace{-3pt}\rotatebox{90}{\makebox[0pt]{MLP}}} & DBA & \underline{1.0809} & 1.0943 & \textbf{1.0779} & 1.3559 & 1.1 & 1.8258 & 1.0997 & 1.3427 \\
 & Jittering & \textbf{1.0662} & 1.0868 & \underline{1.0772} & 1.0828 & 1.0928 & 1.646 & 1.0997 & 1.3427 \\
 & M-Warp & 1.0823 & \underline{1.0821} & \textbf{1.0802} & 1.1153 & 1.1098 & 1.5976 & 1.0997 & 1.3427 \\
 & Scaling & \textbf{1.0571} & 1.064 & \underline{1.0613} & 1.0768 & 1.1038 & 1.6913 & 1.0997 & 1.3427 \\
 & MBB & \textbf{1.072} & \underline{1.0796} & 1.08 & 1.0943 & 1.106 & 1.6498 & 1.0997 & 1.3427 \\
 & TSMixup & \underline{1.071} & \textbf{1.0695} & 1.0773 & 1.1189 & 1.1072 & 1.6795 & 1.0997 & 1.3427 \\
 & T-Warp & \textbf{1.089} & \underline{1.0962} & 1.1031 & 1.2178 & 1.5132 & 1.7219 & 1.0997 & 1.3427 \\
\noalign{\smallskip}
\cline{1-10}
\noalign{\smallskip}
\multirow{7}{*}{\hspace{-3pt}\rotatebox{90}{\makebox[0pt]{NHITS}}} & DBA & \textbf{1.0816} & 1.1048 & \underline{1.0829} & 1.0998 & 1.0919 & 1.7756 & 1.1171 & 1.3427 \\
 & Jittering & 1.0997 & \textbf{1.0773} & 1.1001 & 1.0964 & \underline{1.0927} & 1.6883 & 1.1171 & 1.3427 \\
 & M-Warp & 1.0889 & \textbf{1.0856} & \underline{1.0869} & 1.118 & 1.1233 & 1.6387 & 1.1171 & 1.3427 \\
 & Scaling & \underline{1.0601} & \textbf{1.0566} & 1.067 & 1.098 & 1.1168 & 1.6839 & 1.1171 & 1.3427 \\
 & MBB & 1.0889 & \underline{1.0883} & \textbf{1.0773} & 1.0907 & 1.1301 & 1.682 & 1.1171 & 1.3427 \\
 & TSMixup & \underline{1.0827} & \textbf{1.0817} & 1.0968 & 1.1131 & 1.1116 & 1.6577 & 1.1171 & 1.3427 \\
 & T-Warp & \textbf{1.0852} & 1.1006 & \underline{1.0863} & 1.2205 & 1.5009 & 1.765 & 1.1171 & 1.3427 \\
\noalign{\smallskip}
\cline{1-10}
\noalign{\smallskip}
 & Average & \textbf{1.0835} & \underline{1.0852} & 1.0921 & 1.1355 & 1.1662 & 1.695 & 1.116 & 1.3427 \\
 & Avg. Rank & \textbf{2.67} & \underline{2.79} & 2.97 & 4.3 & 4.4 & 7.5 & 4.02 & 7.34 \\
\noalign{\smallskip}
\cline{1-10}
\bottomrule
\end{tabular}%
}
\end{table}

The results suggest that online augmentation using \texttt{Online} or \texttt{Online(E)} show the best and second-best average MASE and average rank, respectively. The average rank denotes the average relative position of each method (1 being the best) across the 126 variants of dataset, neural network architecture, and time series synthetic generation method. 

The benefits of online data augmentation are consistent across the 3 learning algorithms, and 7 time series data generation methods tested. Notwithstanding, \texttt{Offline(1)} is more competitive when using \texttt{MLP}. When the forecasting models are trained using \texttt{KAN}, online augmentation approaches always outperform offline ones across all time series synthetic data generation methods.

Among the offline approaches, \texttt{Offline(1)} shows the best performance. This suggests that creating one time series for each one in the dataset leads to better forecast accuracy relative to creating 10, as done by Bandara et al. \cite{bandara2021improving} or matching the number of time series created by \texttt{Online} and \texttt{Online(E)}.

Varying the parameters of the time series synthetic data generation methods leads to a comparable forecasting accuracy when using online data augmentation. However, this process severely reduces performance in offline approaches (\texttt{Offline(=,E)}). Except for \texttt{Offline(=,E)}, all approaches outperform \texttt{Naive}. This outcome validates the forecast accuracy of the trained neural networks against a standard baseline. 

\begin{figure}[ht]
    \centering
    \includegraphics[width=\textwidth, trim=0cm 0cm 0cm 0cm, clip=TRUE]{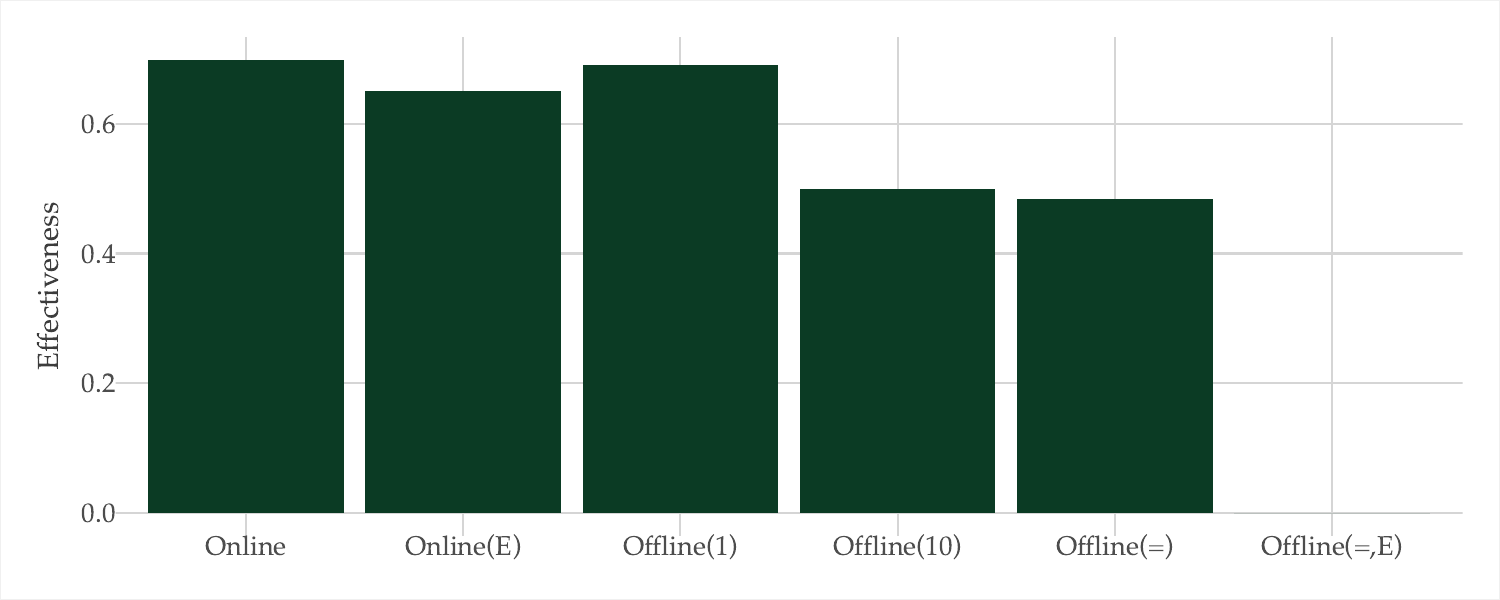}
    \caption{Effectiveness of each data augmentation approach in terms of the ratio of times each one outperforms the \texttt{Original} approach that does not apply data augmentation.}
    \label{fig:effectiveness}
\end{figure}

The average MASE and average rank of \texttt{Online}, \texttt{Online(E)}, and \texttt{Offline(1)} are better than that of \texttt{Original}. We illustrate the effectiveness of each data augmentation method in Figure \ref{fig:effectiveness}, which shows the ratio of times each approach outperforms the \texttt{Original} across the 126 problem variants (6 datasets times 3 neural networks times 7 synthetic data generators). Overall, only \texttt{Online}, \texttt{Online(E)}, and \texttt{Offline(1)} show an effectiveness above 60\%.

\begin{figure}[ht]
    \centering
    \includegraphics[width=\textwidth, trim=0cm 0cm 0cm 0cm, clip=TRUE]{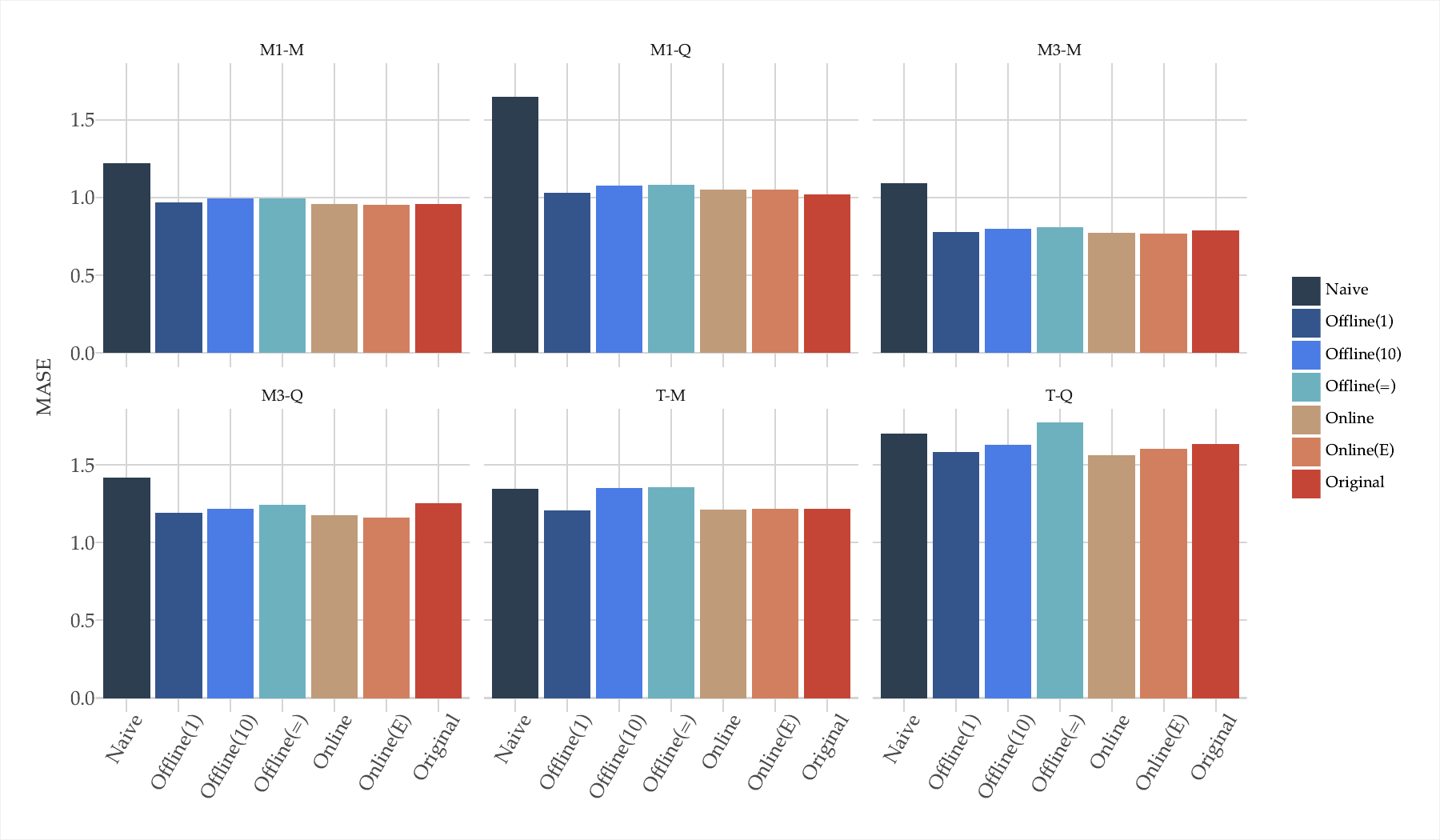}
    \caption{Forecasting accuracy by each dataset and data augmentation approach, averaged across neural network architecture. \texttt{Offline(E,=)} is removed for visualization purposes.}
    \label{fig:mase_by_approach_ds}
\end{figure}

Table~\ref{tab:scores_by_ds} shows the results averaged across the 6 datasets. To get a more nuance view on the results across datasets, we evaluate how the accuracy scores of each data augmentation approach varies by dataset in Figure \ref{fig:mase_by_approach_ds}. \texttt{Online}, \texttt{Online(E)}, and \texttt{Offline(1)} show better forecast accuracy relative to \texttt{Original} across all 6 datasets. However, \texttt{Original} shows a competitive performance in some datasets, such as M1-Q, M1-M, or T-M.

\begin{figure}[ht]
    \centering
    \includegraphics[width=\textwidth, trim=0cm 0cm 0cm 0cm, clip=TRUE]{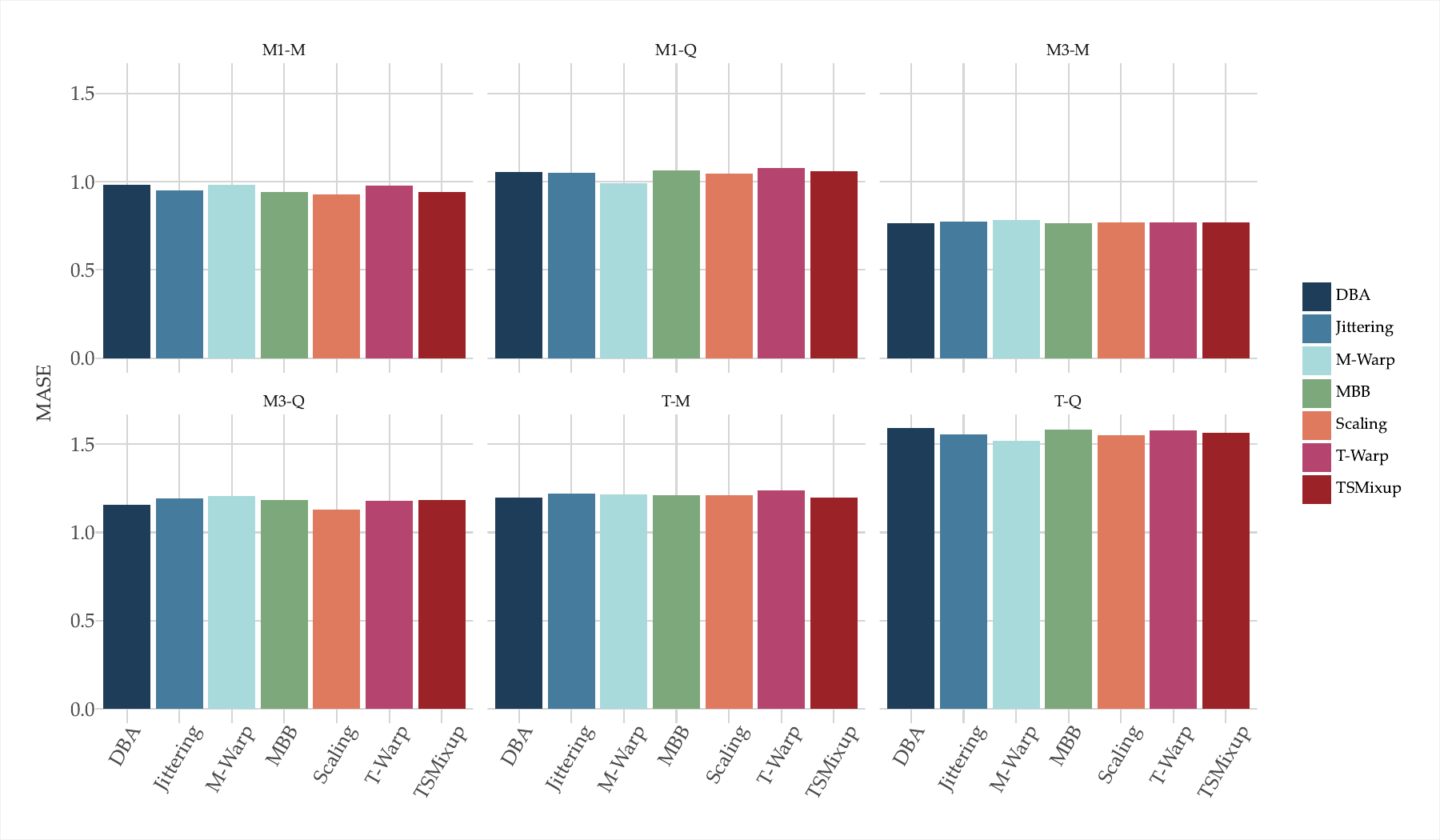}
    \caption{Forecasting accuracy of \texttt{Online} by each dataset and synthetic data generation method, averaged across neural network architecture.}
    \label{fig:mase_by_model_ds}
\end{figure}

In terms of synthetic data generation method, \texttt{Scaling} appears to be the operation that maximizes the overall forecasting accuracy according to Table \ref{tab:scores_by_ds}. We evaluate how the accuracy scores of these approaches vary by dataset in Figure \ref{fig:mase_by_model_ds}. This figure shows the MASE values of \texttt{Online} by each dataset and synthetic data generation method, averaged across neural network architecture. Overall, there is a noticeable variance in relative performance. \texttt{Scaling} shows the best performance in M1-M and M3-Q, while magnitude warping (\texttt{M-Warp}) leads to better MASE in M1-Q and T-Q. In the remaining two datasets, all approaches perform comparably.

\section{Discussion}\label{sec:5}

The main contribution of this work is a method for applying online data augmentation during the training process of neural networks for univariate time series forecasting.  Our extensive empirical analysis provides evidence about the effectiveness of online data augmentation across different datasets, neural network architectures, and synthetic data generation methods. 

First, the systematic superiority of online augmentation across different architectures and datasets suggests that the dynamic nature of synthetic sample generation provides an advantage over offline (before training) augmentation. This may be because online augmentation effectively increases the diversity of training samples while maintaining their relevance to the original data distribution since synthetic samples are coupled with real ones at each batch. 

Our observation that \texttt{Offline(1)} outperforms other offline approaches generating more synthetic samples challenges the idea that more augmented data is always better. This suggests that the quality and timing of synthetic sample generation may be more crucial than quantity. Online augmentation employs this principle with its online generation process that couples synthetic time series with real ones at each training batch.

The architecture-specific variations in improvement indicate that the benefits of online augmentation may be influenced by model capacity and learning dynamics. This interaction between augmentation strategy and model architecture opens up interesting possibilities for joint optimization of both components. In future work, we will study how to improve the data augmentation process during training. For example, introduce a mechanism that weights time series based on their impact on the training process, akin to boosting methods \cite{friedman2002stochastic}. Future work could also explore expanding this approach to multivariate time series and investigating more sophisticated mechanisms for adapting the augmentation process based on model learning dynamics.

\section{Conclusions}\label{sec:6}

Data augmentation techniques are useful in contexts where an insufficient sample size is available to train adequate deep learning models. In univariate time series forecasting problems, data augmentation is typically done offline before the training process. This paper proposes a novel framework for online time series data augmentation to train more accurate neural networks. This framework works by augmenting each batch of time series in each training or validation step. A set of extensive experiments was conducted, suggesting online data augmentation leads to better forecasting accuracy relative to offline approaches or no data augmentation. Improvements were consistent across different neural network architectures and synthetic time series generation techniques. 

Future work could explore adaptive data augmentation processes that evolve during training to maximize performance \cite{hou2023learn}. This could involve dynamically selecting augmentation techniques based on the characteristics of the time series or the model's current performance. Additionally, investigating the combination of multiple augmentation methods and developing new techniques specifically designed for online settings could further improve forecasting accuracy. The framework could also be extended to handle different types of time series, such as those with strong seasonality or unit roots.

\section*{Acknowledgements}

This work was partially funded by projects AISym4Med (101095387) supported by Horizon Europe Cluster 1: Health, ConnectedHealth (n.º 46858), supported by Competitiveness and Internationalisation Operational Programme (POCI) and Lisbon Regional Operational Programme (LISBOA 2020), under the PORTUGAL 2020 Partnership Agreement, through the European Regional Development Fund (ERDF) and NextGenAI - Center for Responsible AI (2022-C05i0102-02), supported by IAPMEI, and also by FCT plurianual funding for 2020-2023 of LIACC (UIDB/00027/2020 UIDP/00027/2020)

\bibliographystyle{spmpsci.bst}






\end{document}